\documentclass[conference]{IEEEtran}
\IEEEoverridecommandlockouts
\usepackage{cite}
\usepackage{amsmath,amssymb,amsfonts}
\usepackage{graphicx}
\usepackage{textcomp}
\usepackage{xcolor}
\usepackage{algorithmicx,algorithm}
\usepackage{booktabs} 
\usepackage{multirow}
\usepackage{amssymb}
\usepackage{longtable}
\usepackage{color}
\usepackage{amsmath}
\usepackage{mathrsfs}

\usepackage{subcaption}
\usepackage{longtable}
\usepackage{ltxtable}
\usepackage{tabularx}
\usepackage{amsmath}
\usepackage{mathrsfs}
\usepackage{multicol}
\usepackage{lipsum}
\usepackage{booktabs}
\usepackage{multirow}
\usepackage{dsfont}
\usepackage{mathbbol}
\usepackage{cleveref}
\usepackage{bbm}
\usepackage{url}
\usepackage{graphicx}
\usepackage{bbding}
\usepackage[normalem]{ulem}

\usepackage{pifont} 
\newcommand{\cmark}{\ding{51}} 
\usepackage{newtxtext,newtxmath} 

\def\BibTeX{{\rm B\kern-.05em{\sc i\kern-.025em b}\kern-.08em
    T\kern-.1667em\lower.7ex\hbox{E}\kern-.125emX}}
\begin{document}

\title{BEACON: Bridging Multimodal Semantics and Asynchronous Propagation for Cloud System Anomaly Detection}


\author{
    \IEEEauthorblockN{
        Zhe Li\textsuperscript{1,2}, Tengxue Zhang\textsuperscript{1,2}, Yang Shu\textsuperscript{1}, Tung Kieu\textsuperscript{3}, Lunting Fan\textsuperscript{2}, Yingying Zhang\textsuperscript{2}$^{*}$, Bin Yang\textsuperscript{1}
    }
    \IEEEauthorblockA{
        \textit{\textsuperscript{1}East China Normal University, \textsuperscript{2}Alibaba Cloud Computing},  \textsuperscript{3}Aalborg University \\
    \{zheli,txzhang\}@stu.ecnu.edu.cn, \{yshu,byang\}@dase.ecnu.edu.cn, \{tungkvt\}@cs.aau.dk,\\
    \{lunting.fan\}@taobao.com, \{congrong.zyy\}@alibaba-inc.com}
    \thanks{$^{*}$Corresponding author.}
}

\maketitle

\begin{abstract}
Modern large-scale cloud systems continuously generate massive and multimodal observability data, such as time-series metrics, log records, and system dependency graphs, which reflect the health state of the system. 
To ensure reliable cloud system operation, anomaly detection models identify abnormal states by analyzing these multimodal observations. However, existing approaches rely on simple fusion strategies and graph learning, ignoring the semantic gap among modalities and the asynchronous spatio-temporal dependencies across system instances over time. 
To this end, we propose BEACON, a framework to Bridge multimodal sEmantics and Asynchronous propagation for Cloud system anOmaly detectioN. 
Specifically, we introduce a Dynamic Interpretable Multimodal Aggregator (DIMA) that leverages an information-bottleneck mechanism to compress information from the semantics and temporal dynamics of metrics, and aligns them with logs in a shared prototype space. 
Furthermore, we propose an Asynchronous Spatio-Temporal Propagator (ASTP) that captures temporal dependencies with multi-scale temporal attention and models spatial relations with topology-aware aggregation. This enables asynchronous propagation of anomalies across system instances. 
Extensive experiments on three real-world cloud system datasets demonstrate that \textbf{BEACON} consistently outperforms state-of-the-art baselines in anomaly detection.

\end{abstract}


\section{Introduction}
Modern large-scale cloud systems (e.g., Alibaba Cloud, AWS) host vast numbers of computing machines that support diverse services for businesses and users~\cite{pan2023MagicScaler,rcrank}. The operational states of these systems are continuously monitored and recorded through multimodal observations, including time-series metrics (e.g., CPU usage, memory utilization), unstructured or semi-structured logs (e.g., textual descriptions of error events), and system dependency graphs (e.g., relationships among system instances). 
System anomalies, such as software bugs, misconfigurations, or sudden traffic spikes, can severely disrupt service availability, propagate across dependent instances, and even trigger large-scale service outages. Therefore, accurate and timely anomaly detection is essential in modern large-scale cloud systems.

\begin{figure}[t]
  \centering
  \includegraphics[width=1\linewidth]{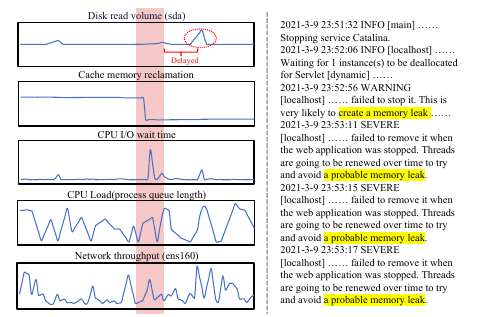}
  \caption{An example of multimodal observations during an anomaly event on the AIOps2021 dataset. The left side presents several system metrics, where the red region indicates the anomalous time period. The right side shows corresponding logs, and the yellow highlights mark anomalous messages.}
  \label{fig: limit}
\end{figure}

Traditional anomaly detection methods primarily rely on predefined rules, expert experience, or basic statistical analysis~\cite{breunig2000lof, goldstein2012histogram, yeh2016matrix}. 
Although these approaches are computationally efficient, they are inherently static and struggle to adapt to the highly dynamic and complex runtime behaviors exhibited by modern large-scale cloud systems. With the advancement of machine learning, single-modal methods emerge to model the temporal dynamics in metrics~\cite{Audibert2020usad, su2019OmniAnomaly, Tuli2022tranad,li2026tiweaver, ZhangOSCGY26} or parse anomalous events from logs~\cite{lee2023hase, zhang2024multilog, xie2025mmfp}. However, these approaches miss anomalies that manifest in other modalities, such as slow jobs caused by task backlog reflected in metrics without corresponding log errors. To address this, recent studies explore jointly modeling logs and metrics by leveraging complementary cross-modal information~\cite{famos,AnoFusion,Medicine}, and further incorporate structural dependencies among system instances through invocation trace analysis or graph learning techniques~\cite{art, eadro}. 
Despite these efforts, these methods suffer from significant limitations: they often adopt simple fusion strategies and assume synchronous dependencies, which fall short in real-world large-scale cloud systems. 
Therefore, building models that can effectively align cross-modal information and capture asynchronous spatio-temporal dependencies presents two challenges:

\textbf{Lack of fine-grained semantic alignment across modalities during multimodal fusion}. 
In large-scale cloud systems, massive volumes of multimodal observations are continuously generated. While these observations provide a comprehensive view of system health, anomaly-relevant information is often buried deep within a ``sea'' of voluminous operational data. For example, a ``\textit{memory leak}'' anomaly may simultaneously cause a spike in ``\textit{CPU I/O wait time}'' and a drop in ``\textit{cache memory reclamation}'', along with a burst of ``\textit{WARNING}'' and ``\textit{SEVERE}'' log events reporting ``\textit{probable memory leak}'', while most other metrics (i.e., ``\textit{network throughput}'') remain normal (as illustrated in Fig.~\ref{fig: limit}). 
However, existing approaches typically align multimodal data by timestamps or apply attention to fuse modality-specific representations at the sequence level, leading to coarse-grained multimodal integration~\cite{famos}. Such modality fusion lacks dynamic temporal interaction within each modality. While some methods attempt fine-grained alignment by extracting event templates~\cite{DiagFusion} or adopting gradient modulation to balance each modality's contribution~\cite{Medicine}, they treat all observations as equally important and cannot distinguish anomaly-relevant information from noise. 
This underscores the need for a semantically guided fusion mechanism that can adaptively identify anomaly-relevant information across modalities while suppressing irrelevant noise.

\textbf{Lack of asynchronous spatio-temporal dependency modeling in cloud systems}.
Instances (e.g., containers or virtual machines) in large-scale cloud systems are interconnected through complex dependencies. In such systems, a fault at one component frequently triggers a ripple effect, propagating along the call chains. Consequently, anomalies may propagate across connected instances in the system graph with varying delays (e.g., an upstream database anomaly causing a downstream API timeout minutes later). 
Existing approaches primarily characterize these spatio-temporal dependencies using graph learning methods~\cite{eadro} or dynamic service call traces~\cite{Guo0WLJDXS20}. Moreover, some methods apply aggregation algorithms over system dependency graphs to the evolution of system states, such as random walks~\cite {ClusterRCA}. 
Despite their effectiveness in modeling spatio-temporal dependencies, these methods do not explicitly capture the temporal delays in anomaly propagation across the system. As illustrated in Fig.~\ref{fig: limit}, increased ``\textit{disk read volume}'' caused by a ``\textit{memory leak}'' may only appear some time after the anomaly begins. 
Spatio-temporal modeling that aligns observations at the same timestamp may introduce misleading correlations. Consequently, modeling asynchronous spatio-temporal dependencies across system instances is important for improving anomaly detection in cloud systems.

To address these challenges, we propose \textbf{BEACON}, a framework to \underline{B}ridge multimodal s\underline{E}mantics and \underline{A}synchronous propagation for \underline{C}loud system an\underline{O}maly detectio\underline{N} via semantic multimodal fusion and asynchronous spatio-temporal modeling.

\textbf{To address the first challenge}, we propose a Dynamic Interpretable Multimodal Aggregator (DIMA). Rather than indiscriminately projecting multimodal observations into a shared latent space, DIMA explicitly leverages fine-grained dynamic semantics to guide the fusion process. Specifically, we first introduce an Information Bottleneck (IB) mechanism that jointly evaluates the intrinsic semantics and the patch-level temporal dynamics of each metric, adaptively filtering out redundant noise while retaining anomaly-relevant information. After selection, highly informative metric embeddings are then fused with log embeddings to form unified multimodal embeddings. 
Furthermore, we design a semantic prototype layer that projects the fused embeddings into an interpretable semantic space spanned by a set of learnable prototypes, endowing each representation with explicit semantic meaning. Through this process, DIMA effectively prevents sparse anomaly information from being overwhelmed by abundant normal noise, yielding highly distinguishable multimodal representations.

\textbf{To address the second challenge}, we design an Asynchronous Spatio-Temporal Propagator (ASTP) that models asynchronous spatio-temporal dependencies across system instances. ASTP first constructs a heterogeneous spatio-temporal graph based on temporal continuity and system dependency graphs. 
To capture asynchronous delays, ASTP introduces a multi-scale temporal attention mechanism. It injects a learnable mixture of time-decay kernels as a distance-aware bias into the attention scores, enabling the model to emphasize relevant historical states across different temporal ranges, modeling asynchronous temporal dependencies.
To capture spatial dependencies, ASTP introduces a topology-aware spatial aggregation module that leverages spatial attention to account for non-uniform influences among neighboring instances.
Through this process, ASTP accurately captures asynchronous spatio-temporal dependencies, improving the reliability of anomaly detection for modern large-scale cloud systems.

\begin{figure*}[t]
  \centering
  \includegraphics[width=1\linewidth]{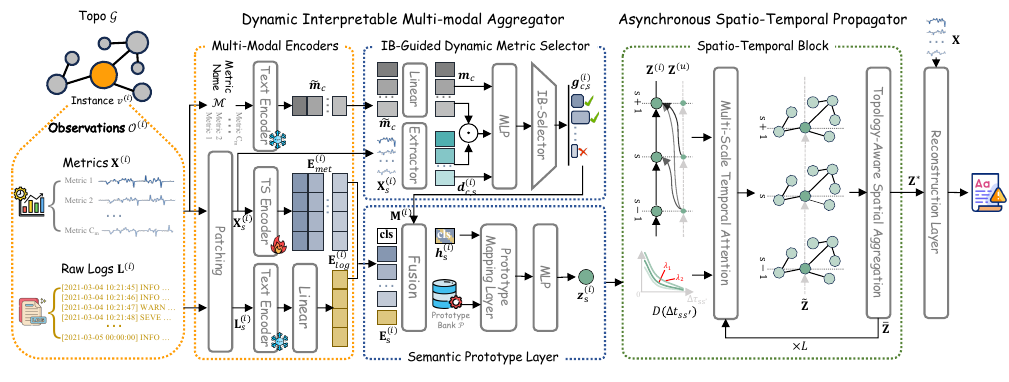}
  \caption{The overall framework of \textbf{BEACON}. On the left, multimodal observations (i.e., metrics, logs, and system dependency graphs) are first encoded into high-dimensional embeddings through modality-specific encoders. DIMA then dynamically distills essential information from massive metrics and aligns metric and log representations in a shared prototype space. Finally, ASTP models asynchronous temporal dependencies and topology-aware spatial interactions across instances to produce representations for anomaly detection.}
  \label{fig: model}
\end{figure*}

In summary, our contributions are as follows:
\begin{itemize}
    \item We propose \textbf{BEACON}, a multimodal anomaly detection framework for cloud systems that detects anomalies by capturing the correlations among multimodal observations, including metrics, logs, and system dependency graphs.
    
    \item We employ Dynamic Interpretable Multimodal Aggregator (DIMA), which leverages fine-grained semantics to guide multimodal fusion, suppressing redundant noise while yielding interpretable and unified representations.
    
    \item We design an Asynchronous Spatio-Temporal Propagator (ASTP) to model multi-scale temporal dependencies along the system dependency graphs, enhancing anomaly detection reliability in modern large-scale cloud systems.

    \item Extensive experiments on three real-world datasets demonstrate that \textbf{BEACON} outperforms the evaluated baselines across the three datasets.

\end{itemize}

\section{Preliminaries}

We formalize the multimodal observations collected from cloud systems and define the anomaly detection task.

\subsection{Multimodal Monitoring Data}
\textit{Definition 1:} \textbf{Metric Sequence.} 
Metrics are numerical measurements (e.g., CPU utilization, memory usage) sampled at regular intervals, which quantify the continuous resource consumption and performance states of the system. Let $\mathcal{M}=\{m_c\}_{c=1}^C$ denote the set of $C$ metric names. Given a time window of length $T$, the metric sequence is defined as $\mathbf{X} \in \mathbb{R}^{T \times C}$.

\textit{Definition 2:} \textbf{Log Sequence.}
Logs are unstructured or semi-structured textual records generated by system applications in an event-driven manner, capturing discrete execution events and exceptions. Over the time window $T$, the raw log stream is first aggregated within each time step and then represented as a sequence $\mathbf{L}=\langle\ell_1,\ell_2,\dots,\ell_T\rangle$.

\textit{Definition 3:} \textbf{System Dependency Graph.}
The cloud system structure is modeled as a stable directed graph $\mathcal{G}=(\mathcal{V}, \mathcal{E})$. Here, $\mathcal{V}=\{v^{(1)}, v^{(2)}, \dots, v^{(N)}\}$ represents the set of $N$ instances deployed in a cloud system, such as containers or virtual machines. $\mathcal{E}$ denotes the dependency relationships among these instances.

\subsection{Problem Formulation}

Given multimodal observations collected from a cloud system, including metrics, logs, and a dependency graph, the anomaly detection task aims to identify which time intervals contain anomalous behaviors. Specifically, for each instance in the system, the model takes the historical multimodal observations within a time window as input and learns representations that capture both cross-modal semantics and spatio-temporal dependencies. Based on these representations, the model detects whether each time interval is normal or anomalous.

To formalize this task, given the dependency graph $\mathcal{G}$ and a time window of length $T$, let $\mathcal{O}=\{\mathcal{O}^{(1)}, \mathcal{O}^{(2)}, \dots, \mathcal{O}^{(N)}\}$ denote the multimodal observations of $N$ instances, where $\mathcal{O}^{(i)}=(\mathbf{X}^{(i)}, \mathbf{L}^{(i)})$ consists of the metric and log sequences of instance $v^{(i)}$. The objective is to learn a mapping function $\mathcal{F}$ parameterized by $\theta$ to detect the health state of the cloud system:

\begin{equation}
\mathcal{F}(\mathcal{G},\mathcal{O};\theta) \rightarrow \hat{\mathcal{Y}}, 
\end{equation}
where $\hat{\mathcal{Y}}=\{(t_{s}^{(i)},t_{e}^{(i)}, y^{(i)})\}_{i=1}^N$ denotes the set of anomalous events of $N$ instances. In each tuple, $[t_{s}^{(i)},t_{e}^{(i)}]$ represents the temporal interval of the anomalous segment for instance $i$, and $y^{(i)} \in \mathcal{A}$ denotes its specific anomaly type label from the predefined category set $\mathcal{A}$ (e.g., network latency, memory leak).

\section{Methodology}

\subsection{Overview}

We propose a multimodal anomaly detection framework, \textbf{BEACON}, to accurately identify anomalies in cloud systems by leveraging multimodal observations (i.e., metrics, logs, and system dependency graphs). The overall architecture of \textbf{BEACON} is illustrated in Fig.~\ref{fig: model}. It consists of two key components: a Dynamic Interpretable Multimodal Aggregator (DIMA) and an Asynchronous Spatio-Temporal Propagator (ASTP).

The Dynamic Interpretable Multimodal Aggregator (DIMA) first employs modality-specific encoders to map the raw multimodal observations into high-dimensional embeddings to enrich information. It then applies an Information Bottleneck (IB) mechanism to dynamically compress massive observations into compact features, which reflect the health state of the cloud system. 
Subsequently, DIMA leverages learnable prototypes to project the compressed metric representations and log embeddings into a shared prototype space, bridging the semantic gap across modalities. 

The Asynchronous Spatio-Temporal Propagator (ASTP) models complex asynchronous correlations by constructing a spatio-temporal graph. Specifically, to capture temporal dynamics, ASTP introduces a learnable mixture of multi-scale time-delay kernels to model patterns across diverse temporal ranges. To capture spatial dependencies, it leverages a graph attention network to strengthen relational interactions among connected instances. Finally, the resulting spatio-temporally enriched multimodal representations are effectively utilized for downstream anomaly detection.

\subsection{Dynamic Interpretable Multimodal Aggregator}
\label{sec: fusion}
Cloud systems generate massive volumes of multimodal observations, making it challenging to identify key information related to anomalies. In addition, the heterogeneity between metrics and logs introduces an inherent modality gap, making it difficult to model them in a unified manner. To address these issues, we propose a Dynamic Interpretable Multimodal Aggregator (DIMA), which distills essential information from the raw observations and projects them into a shared semantic space at the patch level.

\subsubsection{Multimodal Encoders}
To effectively leverage information from multiple modalities, we use modality-specific encoders to extract distinctive features from each modality. Specifically, for a given instance $v^{(i)}$, we divide its observation sequence over a time window $T$ into non-overlapping patches $\langle \mathcal{O}_{1}^{(i)}, \mathcal{O}_{2}^{(i)}, \dots, \mathcal{O}_{S}^{(i)} \rangle$ following the standard patching mechanism in Triformer~\cite{cirstea2022triformer}, where $S=\lceil \frac{T}{W}\rceil $ and $W$ is the predefined patch size. Each patch $\mathcal{O}_{s}^{(i)}=(\mathbf{X}_{s}^{(i)},\mathbf{L}_{s}^{(i)})$ contains metric segment $\mathbf{X}_{s}^{(i)}\in \mathbb{R}^{C \times W}$ and log segment $\mathbf{L}_{s}^{(i)}=\langle \ell_{(s-1)W+1}^{(i)},\dots,\ell_{sW}^{(i)}\rangle$ spanning $W$ time steps.

For a given patch $\mathcal{O}_{s}^{(i)}$, its metrics $\mathbf{X}_{s}^{(i)}$ are projected into $D$-dimensional embeddings via a learnable time series encoder: 
\begin{equation}
    \label{eq2}
    \mathbf{E}_{\text{met},s}^{(i)} = \text{TSEncoder}(\mathbf{X}_{s}^{(i)})=\text{MLP}_{\text{met}} \left(\mathbf{X}_{s}^{(i)}\right)  \in \mathbb{R}^{C \times D}.
\end{equation}
To derive compact semantic descriptors from unstructured or semi-structured and voluminous raw logs, the log segment $\mathbf{L}_{s}^{(i)}$ is first parsed into event templates by Drain~\cite{Drain} with a fixed-depth tree:
\begin{equation}
    \mathcal{T}_{s}^{(i)}=\text{Drain}(\mathbf{L}_{s}^{(i)})=\{(\tau_r,f_{s,r}^{(i)})\}_{r=1}^{R},
\end{equation}
where $\tau_r$ is the $r$-th event template, and $f_{s,r}^{(i)}$ denotes the occurrence frequency of event $r$. The template set is then converted into a single textual description (e.g., ``[NullPointerException] ×12; [Connection timeout] ×4; ...''), containing both the semantic content and the relative intensity of log events. This description is mapped to a semantic embedding through a frozen pre-trained text encoder:
\begin{equation}
    \tilde{\mathbf{E}}_{\text{log},s}^{(i)}=\text{TextEncoder}(\mathcal{T}_{s}^{(i)}) \in \mathbb{R}^{D_{\text{text}}}.
\end{equation}
Since the text encoder output dimension $D_{\text{text}}$ differs from the metric embedding dimension $D$, a lightweight linear layer is applied to align them:
\begin{equation}
    \label{eq5}
    \mathbf{E}_{\text{log},s}^{(i)} = \text{Linear(}\tilde{\mathbf{E}}_{\text{log},s}^{(i)}) \in \mathbb{R}^{D}.
\end{equation}

\subsubsection{IB-Guided Dynamic Metric Selector}
\label{sec:ib-selection}
Due to the complex design of cloud systems, massive metrics are often noisy and contain redundant information, making it difficult to accurately model the true state of the cloud system. To address this issue, we introduce an Information Bottleneck (IB) mechanism~\cite{ib} that dynamically distills essential information by jointly considering the intrinsic semantics and the patch-level temporal dynamics of each metric, producing compact representations that better capture the system state.

Metrics reflect system states that evolve over time, requiring joint modeling of both their intrinsic semantics and temporal dynamics. Let $y^{(i)}_s \in \mathcal{A} \ \cup \{\text{``Normal''}\}$ denote the state of the instance $v^{(i)}$ during the $s$-th patch, the objective is to learn a compressed representation $\boldsymbol{z}_{s}^{(i)}$ that maximally preserves information of $y^{(i)}_s$, while discarding redundant noise from the raw metric space: 
\begin{equation}
    \label{eq:ib-objective}
    \max_{\boldsymbol{z}_{s}^{(i)}} I(\boldsymbol{z}_{s}^{(i)}; y^{(i)}_s) - \beta I(\boldsymbol{z}_{s}^{(i)};\mathbf{X}_{s}^{(i)},\mathcal{M}),
\end{equation}
the $\beta\!>\!0$ controls the compression rate. 

For the semantic prior, the textual metric name $m_c$ is encoded by the $\text{TextEncoder}(\cdot)$ to obtain the metric name embedding $\boldsymbol{m}_c$:
\begin{equation}
    \label{eq7}
    \boldsymbol{m}_c = \text{Linear}( \text{TextEncoder}(m_c))\in \mathbb{R}^{D}.
\end{equation}
Then, we extract standard statistical features to capture temporal dynamics, including mean $\mu$, variance $\sigma$, skewness $\gamma_1$, kurtosis $\gamma_2$, maximum first-order difference $\max|\Delta|$, and coefficient of variation $\sigma_{\Delta}/(|\mu|+\epsilon)$. These dynamic features are processed by an MLP to obtain the temporal embedding:
\begin{equation}
    \label{eq8}
    \boldsymbol{d}_{c,s}^{(i)} = \text{MLP}_{\text{dyn}}\big[\,\mu,\,\sigma,\,\gamma_1,\,\gamma_2,\;\max|\Delta| ,\,\sigma_{\Delta}/(|\mu|+\epsilon)\big].
\end{equation}
These two embeddings are fused to produce compact representations that focus on informative metrics. Specifically, we first compute a scalar logit:
\begin{equation}
    \label{eq9}
    l_{c,s}^{(i)} =
    \mathrm{MLP}_{\text{fus}}(\big[\boldsymbol{m}_c;\boldsymbol{d}_{c,s}^{(i)};\boldsymbol{m}_c\odot\boldsymbol{d}_{c,s}^{(i)}\big])\in\mathbb{R}\ .
\end{equation}
where the $\odot$ denotes Hadamard product. The retention probability of the $c$-th metric is computed as $p_{c,s}^{(i)}=\sigma(l_{c,s}^{(i)})$, where $\sigma(\cdot)$ denotes the standard sigmoid activation function that maps the logit into (0,1). A binary gating variable $g_{c,s}^{(i)} \sim \text{Bernoulli}(p_{c,s}^{(i)})$ is sampled to control whether the metric is retained. 

The raw metric embeddings $\mathbf{E}_{\text{met},s}^{(i)}$ are filtered by the learned gates $\mathbf{G}_{s}^{(i)}=\{ g_{c,s}^{(i)}\}_{c=1}^{C}$ to produce compact metric embeddings that focus on informative information:
\begin{equation}
    \label{eq10}
    \hat{\mathbf{E}}_{\text{met},s}^{(i)} = \mathbf{G}_{s}^{(i)} \cdot \mathbf{E}_{\text{met},s}^{(i)} \in \mathbb{R}^{C \times D}.
\end{equation}
To realize the compression term in~\eqref{eq:ib-objective}, we define the prior distribution $\text{Bernoulli}(\pi)$ following the previous work~\cite{MiaoLL22}, where $\pi \in [0,1]$ is a prior retention rate. For each instance $i$ at the $s$-th patch, we calculate Kullback-Leibler (KL) divergence as the tractable loss to constrain the retention of redundant metrics:
\begin{equation}
    \label{eq11}
    \mathcal{L}_{\text{kl},s}^{(i)} = \frac{1}{C}\sum_{c=1}^{C} D_{KL} \Big( \text{Bernoulli}(p_{c,s}^{(i)}) \ \big|\ \text{Bernoulli}(\pi) \Big).
\end{equation}

\subsubsection{Semantic Prototype Layer}
\label{sec:proto}

To bridge the semantic gap between the compact metric and log embeddings, we introduce learnable prototypes as semantic anchors, mapping heterogeneous modalities into a unified, interpretable semantic space. Specifically, we employ a self-attention to capture interactions between metrics and logs:
\begin{equation}
    \label{eq12}
    \text{Self-Attn}(\mathbf{E}_s^{(i)}) =  \mathbf{M}_s^{(i)}\cdot\text{Softmax}\left( \frac{\mathbf{Q}_s^{(i)} \mathbf{K}_s^{(i)\top}}{\sqrt{D}} \right) \mathbf{V}_s^{(i)},
\end{equation}
where the query, key, and value matrices $\mathbf{Q}_s^{(i)}$, $\mathbf{K}_s^{(i)}$, and $\mathbf{V}_s^{(i)}$ are projected from the multimodal sequence $\mathbf{E}_s^{(i)}=\big[\,\langle \text{cls}\rangle;\hat{\mathbf{E}}_{\text{met},s}^{(i)};{\mathbf{E}}_{\text{log},s}^{(i)}\ \big]$, in which $\langle \text{cls}\rangle$ is a learnable token used to aggregate the multimodal information. The binary vector $\mathbf{M}_s^{(i)}=\{1\}\cup\mathbf{G}_{s}^{(i)}\cup\{1\} \in \mathbb{R}^{2+C}$ serves as the attention mask to disable interactions for suppressed metrics. We take the hidden state corresponding to the $\langle \text{cls} \rangle$ token as the fused multimodal representation $\boldsymbol{h}_s^{(i)}\!\in\!\mathbb{R}^{D}$.

Let $\mathcal{P}=\{\boldsymbol{p}_k\}_{k=1}^{K}\!\in\!\mathbb{R}^{K\times D}$ denotes a bank of learnable prototypes that act as semantic anchors in the shared representation space. Each prototype represents a typical joint pattern of metrics and logs (e.g., a surge in CPU utilization accompanied by program startup logs). The prototype bank is initialized through an LLM-guided procedure (detailed in Section~\ref{sec: proto_init}).

Given the fused representation $\boldsymbol{h}_s^{(i)}$, we compute its similarity with each prototype to obtain the activation scores: 
\begin{equation}
    \label{eq13}
    \alpha_{s,k}^{(i)} = \frac{\exp\!\big(\cos(\boldsymbol{h}_s^{(i)},\,\boldsymbol{p}_k)/\tau_{proto}\big)}{\sum_{k=1}^{K}\exp\!\big(\cos(\boldsymbol{h}_s^{(i)},\,\boldsymbol{p}_{k})/\tau_{proto}\big)} ,
\end{equation}
where $\tau_{proto}$ is a temperature parameter, controlling the sharpness of the similarity distribution. 
We then project the fused representation into the prototype semantic space by aggregating the prototypes according to these activation scores:
\begin{equation}
    \boldsymbol{h}_{\text{proto},s}^{(i)} = \sum_{k=1}^{K} \alpha_{s,k}^{(i)} \cdot\boldsymbol{p}_k \in \mathbb{R}^{D}.
\end{equation}
By expressing multimodal observations as combinations of shared prototypes, heterogeneous metric and log features are effectively mapped into a unified semantic space. 
Finally, we concatenate the two representations and feed the result into an MLP to obtain the final multimodal fusion representation:
\begin{equation}
    \label{eq15}
    \boldsymbol{z}_s^{(i)} = \text{MLP}_{\text{proto}}(\big[\,\boldsymbol{h}_s^{(i)};\,\boldsymbol{h}_{\text{proto},s}^{(i)}\,\big]) \in \mathbb{R}^{D}.
\end{equation}
The resulting representation $\boldsymbol{z}_s^{(i)}$ is served as the initial node feature for the asynchronous spatio-temporal propagator and used to compute the classification loss:

\begin{equation}
    \mathcal{L}_{\text{cls}}=
    \mathbb{E}{(\boldsymbol{z}_{s}^{(i)}, y_s)}
    \left[
    \log q_{\theta}(y_s^{(i)} \mid \boldsymbol{z}_{s}^{(i)})
    \right].
\end{equation}

The overall selection loss is formulated by combining the KL-divergence loss and the classification loss as $\mathcal{L}_{\text{select}}=\mathcal{L}_{\text{kl}}+\mathcal{L}_{\text{cls}}$. Furthermore, to maintain the semantic integrity of the prototype bank, we regularize $\mathcal{P}$ with three complementary objectives: sparsity, diversity, and semantic alignment. The regularization terms are defined as:
\begin{equation}
    \label{eq16}
    \begin{gathered}
    \mathcal{L}_{\text{sparse}} = \frac{1}{NS}\sum_{i=1}^N\sum_{s=1}^S (1-\max_{1\le k\le K}\alpha_{k,s}^{(i)}), \\[2pt]
    \mathcal{L}_{\text{div}}    = \frac{2}{K(K-1)}\sum_{1\le k<k'\le K}\max\!\big(0,\,\cos(\boldsymbol{p}_k,\boldsymbol{p}_{k'})-\delta_{\text{proto}}\big), \\[2pt]
    \mathcal{L}_{\text{align}}  = -\frac{1}{K}\sum_{k=1}^{K}\cos\!\big(\boldsymbol{p}_k,\,\boldsymbol{p}_k^{(\text{text})}\big).
    \end{gathered}
\end{equation}
Here, $\mathcal{L}_{\text{sparse}}$ encourages each patch representation to concentrate on a small number of prototypes, $\mathcal{L}_{\text{div}}$ promotes diversity among prototypes by penalizing highly similar prototype pairs, where $\delta_{\text{proto}}$ is the similarity threshold, and $\mathcal{L}_{\text{align}}$ preserves the initial semantic meaning of prototypes by aligning them with their textual embeddings $\boldsymbol{p}_k^{(\text{text})}$ obtained during initialization. The overall prototype regularizer is given by $\mathcal{L}_{\text{proto}}=\mathcal{L}_{\text{sparse}}+\mathcal{L}_{\text{div}}+\mathcal{L}_{\text{align}}$.

\subsection{Asynchronous Spatio-Temporal Propagator}
\label{sec:stg}
In cloud systems, anomalies rarely remain confined to a single instance. Anomalies in one instance can propagate to related instances with a time lag. To capture such asynchronous dependencies across instances, we propose an Asynchronous Spatio-Temporal Propagator that combines topology-aware message passing with multi-scale temporal modeling.

\subsubsection{Spatio-Temporal Graph Construction}
\label{sec:stg-construction}

Given the system dependencies graph $\mathcal{G}=\{\mathcal{V},\mathcal{E}\}$, we construct a heterogeneous spatio-temporal graph $\mathcal{G}_{\text{st}} = (\mathcal{V}_{\text{st}}, \mathcal{E}_{\text{st}})$ over patch windows. The node set $\mathcal{V}_{\text{st}} = \{v_{s}^{(i)} \mid i \in [1,N], s \in [1, S]\}$ expands each instance $v^{(i)} \in \mathcal{V}$  across patch windows, where $v_{s}^{(i)}$ denotes instance $v^{(i)}$ at the $s$-th patch. Each node is initialized with a multimodal representation $\boldsymbol{z}_s^{(i)} $ of the corresponding patch. The edge set $\mathcal{E}_{\text{st}}$ includes temporal and spatial edges to capture dependencies between nodes: 
\textit{Temporal edges} $\mathcal{E}_{\text{tem}}=\{(v_{s'}^{(u)}, v_{s}^{(i)}) \mid (u,i)\in \mathcal{E}\cup \{(i,i)\},s-s^{*}<s'<s\}$ connect each node to the previous-patch nodes of itself and its neighbors $\mathcal{N}(v^{(i)})$ in $\mathcal{G}$, where $s^*$ denotes the maximum temporal range used to construct temporal edges. 
\textit{Spatial edges} $\mathcal{E}_{\text{spa}}=\{(v^{(u)}_{s},v^{(i)}_{s})\mid (u,i)\in \mathcal{E}\}$ are based on the system dependencies graph $\mathcal{G}$, capturing the spatial interactions among instances at the current patch.

\subsubsection{Multi-Scale Temporal Attention}
\label{sec:stg-temporal}

Anomalies in distributed cloud systems often propagate across instances with varying time delays, caused by factors such as deferred project scheduling and multi-hop dependency chains. To capture such asynchronous temporal dependencies, we introduce a Multi-Scale Temporal Attention mechanism that models delayed interactions across different time horizons using learnable time-decay kernels.

Specifically, let $\Delta t_{ss'} = s - s'$ denote the time interval between $v_s^{(i)}$ and its temporal neighbor $v_{s'}^{(u)} \in \mathcal{N}_{\text{tem}}(v_s^{(i)})$. We model a multi-scale temporal factor $D(\Delta t_{ss'})$ as a mixture of $M$ exponential decays with different learnable decay rates $\lambda_m$:
\begin{equation}
\label{eq17}
D(\Delta t_{ss'}) = \sum_{m=1}^{M} w_m \exp(-\lambda_m \cdot \Delta t_{ss'}),
\end{equation}
where $w_m$ denotes the learnable scalar weight for the $m$-th scale. We incorporate the factor as a temporal bias in attention scores to modulate the influence of historical states with different delays:
\begin{equation}
\label{eq18}
s_{ss'}^{(ui)} = \left( \frac{(\mathbf{W}_Q \boldsymbol{z}_{s}^{(i)})^\top (\mathbf{W}_K\boldsymbol{z}_{s'}^{(u)})}{\sqrt{D}} \right) + D(\Delta t_{ss'}),
\end{equation}
Here, $\mathbf{W}_Q, \mathbf{W}_K \in \mathbb{R}^{D \times D}$ are the query and key projection matrices.
We normalize the scores over the temporal neighborhood using softmax to obtain attention weights $a_{ss'}^{(ui)} = \text{Softmax}(s_{ss'}^{(ui)})$. The temporal representation is then updated as:
\begin{equation}
\label{eq19}
\tilde{\boldsymbol{z}}_{s}^{(i)} = \sum_{v_{s'}^{(u)} \in \mathcal{N}_{\text{tem}}(v_{s}^{(i)})} a_{ss'}^{(ui)}\boldsymbol{z}_{s'}^{(u)}.
\end{equation}

\subsubsection{Topology-Aware Spatial Aggregation}
\label{sec:stg-spatial}
Building upon the temporally updated representations, we further model spatial dependencies among adjacent instances within the same patch using a spatial attention mechanism. Specifically, in the $s$-th patch, we compute the spatial attention score between an instance $v_s^{(i)}$ and its spatial neighbor $v_s^{(u)} \in \mathcal{N}(v_s^{(i)})$ as:
\begin{equation}
\label{eq20}
    s_{s}^{(ui)} = \boldsymbol{a}_{\text{spa}}^\top \text{LeakyReLU} \left( \mathbf{W}_{\text{spa}} [\tilde{\boldsymbol{z}}_{s}^{(u)} \Vert \tilde{\boldsymbol{z}}_{s}^{(i)}] \right),
\end{equation}
where $\mathbf{W}_{\text{spa}} \in \mathbb{R}^{D \times 2D}$ is the feature transformation matrix, $\mathbf{a}_{\text{spa}} \in \mathbb{R}^{D}$ is a learnable weight vector, and $\Vert$ denotes concatenation operation. 
The scores $s_{s}^{(ui)}$ are normalized with a softmax function over all spatial neighbors to get the attention weight $a_{s}^{(ui)}$:
\begin{equation}
    a_{s}^{(ui)} = \frac{\exp(s_{s}^{(ui)})}{\sum_{v_{s}^{(v)} \in \mathcal{N}(v_s^{(i)})} \exp(s_{s}^{(vi)})}.
    \label{eq:spatial_softmax}
\end{equation}
Finally, the spatial representation is updated as:
\begin{equation}
    \bar{\boldsymbol{z}}_{s}^{(i)} = \sigma \left( \sum_{v_{s}^{(u)} \in \mathcal{N}(v_s^{(i)})} a_{s}^{(ui)} \mathbf{W}_{V} \tilde{\boldsymbol{z}}_{s}^u \right),
\label{eq22}
\end{equation}
where $\mathbf{W}_{V} \in \mathbb{R}^{D \times D}$ is the spatial feature projection matrix, $\sigma$ denotes a non-linear activation function, and $\bar{\boldsymbol{z}}_{s}^{(i)}$ represents the final output of the spatio-temporal block for instance $v^{(i)}$ at the $s$-th patch.

To model asynchronous spatio-temporal dependencies across the system, we implement the ASTP as a stack of $L$ Spatio-Temporal Blocks. Each block combines multi-scale temporal attention with topology-aware spatial aggregation, updating node representations by integrating information from historical states and neighbors. The output of the $l$-th block is used as the input to the $(l+1)$-th block. We omit the layer index $(l)$ in the preceding formulations for notational simplicity. After $L$ layers, we obtain the final multimodal spatio-temporal representations of all instances, denoted as $\mathbf{Z}^* \in \mathbb{R}^{N \times S \times D}$.

\begin{algorithm}[t]
\caption{Pseudo-code of \textbf{BEACON} }
\small
\begin{flushleft}
{\bf Input:} 
Multimodal observations $\mathcal{O}$ and a system dependency graph $\mathcal{G}$.

\end{flushleft}

\begin{flushleft}
{\bf Warm-up and Representation Collection:}
\begin{algorithmic}[1]
\State {\bf for} $i=1$ to $N$ {\bf do}
\State \hspace{0.1in} Divide $\mathcal{O}^{(i)}$ into patches $\langle\mathcal{O}_{s}^{(i)}\rangle_{s=1}^S$;
\State \hspace{0.1in} {\bf for} $s=1$ to $S$ {\bf do}
\State \hspace{0.1in} \hspace{0.1in} Encode $\mathcal{O}_s^{(i)}$ into $\mathbf{E}_{\text{met},s}^{(i)}$ and $\mathbf{E}_{\text{log},s}^{(i)}$ using~\eqref{eq2}--\eqref{eq5}; 
\State \hspace{0.1in} \hspace{0.1in} Compute the binary gates $\mathbf{G}_s^{(i)}$ using~\eqref{eq7}--\eqref{eq9}; 
\State \hspace{0.1in} \hspace{0.1in} Filter metrics $\hat{\mathbf{E}}_{\text{met},s}^{(i)}$ by~\eqref{eq10}; 
\State \hspace{0.1in} \hspace{0.1in} Fusion multimodal to get $\boldsymbol{h}_s^{(i)}$ using~\eqref{eq12}; 
\State \hspace{0.1in} {\bf end for}
\State {\bf end for}
\State Update \textbf{BEACON} based on~\eqref{eq11} and~\eqref{eq16}; 
\State Apply K-Means to $\boldsymbol{h}_s^{(i)}$ to obtain the typical representation set $\tilde{\mathcal{P}}$;
\State {\bf return} $\tilde{\mathcal{P}}$;
\end{algorithmic}
\end{flushleft}

\begin{flushleft}
{\bf Prototype Initialization and Joint Optimization:}
\begin{algorithmic}[1]
\setcounter{ALG@line}{12}
\State Annotate $\tilde{\mathcal{P}}$ with $\{\boldsymbol{p}_k^{(\text{text})}\}_{k=1}^K$ using a LLM to construct the prototype bank $\mathcal{P}$;

\State {\bf for} $i=1$ to $N$ {\bf do}
\State Execute Line 1 of the algorithm;
\State \hspace{0.1in} {\bf for} $s=1$ to $S$ {\bf do}
\State \hspace{0.1in} \hspace{0.1in} Execute Lines 4–7 to get the fusion representations $\boldsymbol{h}_s^{(i)}$;
\State \hspace{0.1in} \hspace{0.1in} Map $\boldsymbol{h}_s^{(i)}$ into prototype space $\boldsymbol{z}_s^{(i)}$ on~\eqref{eq13}--\eqref{eq15};
\State \hspace{0.1in} \hspace{0.1in} Construct staptio-temporal graph $\mathcal{G}_{\text{st}}$;
\State \hspace{0.1in} \hspace{0.1in} Compute the multi-scale temporal bias $D(\Delta t_{ss'})$ on \eqref{eq17};
\State \hspace{0.1in} \hspace{0.1in} Model temporal dependencies $\tilde{\boldsymbol{z}}_{s}^{(i)}$ using~\eqref{eq18}--\eqref{eq19};
\State \hspace{0.1in} \hspace{0.1in} Model spatial dependencies $\bar{\boldsymbol{z}}_{s}^{(i)}$ using~\eqref{eq20}--\eqref{eq22};
\State \hspace{0.1in} {\bf end for}
\State {\bf end for}
\State Reconstruct metric sequences $\hat{\mathbf{X}}$ from~\eqref{eq23}--\eqref{eq24};
\State Update \textbf{BEACON} based on~\eqref{eq27}; 

\State {\bf return};
\end{algorithmic}
\end{flushleft}

\label{beacon}
\end{algorithm}

\subsection{Detection and Optimization}
\label{sec:optimization}

\subsubsection{Reconstruction-based Anomaly Detection}
\label{sec: loss}
The spatio-temporal representations $\mathbf{Z}^*$ encode fused multimodal information in a shared semantic space. To detect anomalies, we reconstruct the original metric observations from these representations and measure the deviation between the reconstructed and observed values.

Specifically, for each instance $v^{(i)}$, we first expand the fused representations $\mathbf{Z}^* \in \mathbb{R}^{N \times S \times D}$ into $C$ metric-wise representations using an MLP:
\begin{equation}
    \label{eq23}
    \mathbf{H} = \text{MLP}_{\text{expend}}(\mathbf{Z}^*) \in \mathbb{R}^{N \times C \times S \times D}.
\end{equation}
Then, another MLP maps the metric-wise representations back to the original temporal resolution:
\begin{equation}
    \label{eq24}
    \hat{\mathbf{X}} = \text{MLP}_{\text{recon}}(\mathbf{H}) \in \mathbb{R}^{N \times C \times T}.
\end{equation}

During the training phase, the primary objective is to minimize the reconstruction deviation over normal data. The reconstruction loss $\mathcal{L}_{\text{recon}}$ is formally defined as the average Mean Squared Error (MSE) over all instances:
\begin{equation}
\label{eq25}
\mathcal{L}_{\text{recon}} = \Vert \mathbf{X} - \hat{\mathbf{X}} \Vert_2^2.
\end{equation}

During the inference phase, the metric-wise anomaly score at the $s$-th patch is defined as the reconstructed error $\epsilon_{c,s}^{(i)}$ compared with the original values. For the instance-level anomaly score, we aggregate the metric-wise scores weighted by their dynamic selection probabilities $p_{c,s}^{(i)}$ derived in Section~\ref{sec:ib-selection}:
\begin{equation}
\label{eq26}
\epsilon_{s}^{(i)} = \sum_{c=1}^{C} p_{c,s}^{(i)} \cdot \epsilon_{c,s}^{(i)}.
\end{equation}
We compute an anomaly threshold $\eta$ on the validation set to identify time intervals when an instance becomes anomalous, i.e., $\epsilon_{s}^{(i)} > \eta$. The duration, severity, and scope of instance-level anomalies can then be used to assess the overall health status of the cloud system.

\begin{figure}[t]
  \centering
  \includegraphics[width=1\linewidth]{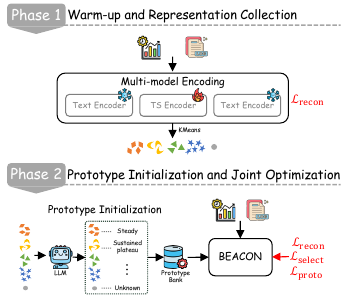}
  \caption{Overview of the two-phase training pipeline.}
  \label{fig: train}
\end{figure}

\subsubsection{Two-Phase Training and Prototype Initialization}
\label{sec: proto_init}
Standard end-to-end optimization of prototypes often suffers from mode collapse and semantic ambiguity. To ensure the prototypes bank captures meaningful, human-readable multimodal information, we design a two-phase training pipeline with an LLM-guided initialization strategy. The overall procedure is summarized in Algorithm~\ref{beacon} and illustrated in Fig.~\ref{fig: train}.

\emph{Phase 1: Warm-up and Representation Collection.}  
We first warm up the network using the reconstruction loss $\mathcal{L}_{\text{recon}}$ to stabilize the multimodal fusion space. We then sample a diverse set of fused patch-level representations $\boldsymbol{h}_s^{(i)}$ from the training set. K-Means~\cite{kmeans} clustering is applied to identify $K-1$ typical joint patterns of metrics and logs, and the cluster centers are used as the patch-level representation set $\tilde{\mathcal{P}}$. To account for unseen patterns, the $K$-th representation is reserved as an unknown state and initialized using the mean representation of samples located in low-density regions.

\emph{Phase 2: Prototype Initialization and Joint Optimization.} For each of the $K-1$ discovered clusters, we sample several nearest neighbors and feed their raw metric sequences and textual logs into an LLM. Guided by a diagnostic prompt, the LLM generates concise semantic labels, which are then projected into the embedding space as textual anchors $\{\boldsymbol{p}_k^{(\text{text})}\}_{k=1}^K$. These textual anchors together with $\tilde{\mathcal{P}}$ constitute the prototype bank $\mathcal{P}$. After the prototype initialization, the model is optimized under the following objective using gradient descent with backpropagation:
\begin{equation}
\label{eq27}
\mathcal{L}_{\text{total}} = \mathcal{L}_{\text{recon}} + \mathcal{L}_{\text{select}} + \mathcal{L}_{\text{proto}}.
\end{equation}

\section{Experiments}

\subsection{Experimental Setup}
\subsubsection{Datasets}
To evaluate the performance of \textbf{BEACON}, we conduct extensive experiments on three cloud system datasets: AIOps2021, Train-Ticket, and D1 (collected from an anonymized production environment of a large-scale cloud system). Table~\ref{tab: Datasets} presents detailed statistics of the three datasets.

\begin{table}[t]
\centering
\caption{The statistics of evaluation datasets.}
\label{tab: Datasets}
\begin{tabular}{c|c|c|c}
\toprule
\textbf{Dataset} & \textbf{AIOps2021} & \textbf{Train-Ticket} & \textbf{D1} \\
\midrule
Observation & 14400 & 11165 & 24689  \\
Instance & 18 & 94 & 14 \\
Frequency & 1 min & 5 seconds & 1 min \\
Feature & 6-90 & 38 & 5-36 \\
Anomaly types & 6 & 12 & 4 \\
Split & 4:2:3 & 2:1:1 & 5:2:3 \\
\bottomrule
\end{tabular}
\end{table}

\begin{itemize}
    \item \textbf{AIOps2021}\footnote{https://aiops-challenge.com} is collected from the management system of a large commercial bank and contains monitoring data from 18 instances (i.e., microservices, servers, databases, and Docker containers). The dataset covers the period from January 2021 to June 2021. Failure records are annotated with several fault types, including Memory, Network Fault, JVM-CPU, JVM-Memory, Disk, and CPU. The labeling process is conducted by two experienced operators independently and cross-checked to ensure consistency. 
    \item \textbf{Train-Ticket}\footnote{https://modelscope.cn/datasets/dcm552/FAMOS-dataset} is collected from the Train-Ticket system for railway ticketing services, supporting operations such as ticket booking, payment, and rescheduling. FAMOS~\cite{famos} uses Locust to inject traffic to simulate user requests and employs ChaosBlade to inject anomalies into the system. The data is collected during the last week of June 2024 and includes monitoring data at the host, JVM, and pod levels. In total, 12 types of anomalies are covered, such as CPU Overload, Network Packet Loss, and Service Latency.
    \item \textbf{D1} is collected from an anonymized production environment of a large-scale cloud system under approved internal research procedures and fully anonymized to protect proprietary and sensitive information. To capture realistic failure scenarios while ensuring system stability, the anomalies are accumulated through routine operational telemetry combined with controlled fault-injection exercises and chaos engineering drills. The dataset contains monitoring data from 14 instances and includes four types of anomalies: Failed Requests, CPU Usage Anomaly, Request Backlog, and Service Unavailability Events. These anomaly categories are labeling classes for benchmarking detection methods and do not reflect the service quality or availability of any production system.
\end{itemize}

\subsubsection{Baselines}
We compare \textbf{BEACON} with several state-of-the-art baselines for cloud system anomaly detection, including both single-modal and multimodal methods. 
Table~\ref{tab: main} summarizes the modalities used by each method.

\begin{itemize}
    \item ContraAD~\cite{ContraAD} is a contrastive learning-based anomaly detection method designed for Flink systems. It constructs positive and negative sample pairs from metric time series and applies cross-contrastive learning to learn discriminative representations, enabling the model to distinguish between normal and anomalous patterns without labels.
    \item TranAD~\cite{Tuli2022tranad} is a transformer-based anomaly detection method for time series. It employs a deep transformer encoder-decoder architecture combined with an adversarial training procedure and a self-conditioning mechanism to amplify reconstruction errors for anomalous points.
    \item Hades~\cite{lee2023hase} is a semi-supervised anomaly detection method for software systems that handles massive logs. It designs a cross-modal attention mechanism to capture the semantic relationships between different logs, allowing the model to detect anomalies by learning from both labeled and unlabeled log sequences. 
    \item MultiLog~\cite{zhang2024multilog} is an anomaly detection method specifically designed for distributed database systems. It extracts multivariate features from logs, including log event counts, semantic embeddings, and temporal patterns, and combines them to identify anomalous behaviors in complex distributed environments.
    \item DiagFusion~\cite{DiagFusion} is a multimodal method that fuses all modalities during data preparation by consolidating them into an event sequence. These events are then encoded with two topology-adaptive graphs for anomaly detection.
    \item Medicine~\cite{Medicine} is a modality-independent anomaly detection framework that identifies anomaly types based on multimodal observations. To preserve the unique characteristics of each modality, it encodes them separately and further adopts multimodal adaptive optimization to balance the learning progress.
    \item FAMOS~\cite{famos} is an anomaly detection method for cloud systems that achieves effective multimodal data fusion. It designs modality-specific encoders to extract features from different observations and employs a fusion module to combine these features.
    \item Eadro~\cite{eadro} presents a multimodal failure diagnosis framework. It designs separate feature extractors for three modalities and then fuses the multimodal features. Eadro learns the dependency-aware status of the system using graph attention networks.
    \item ART~\cite{art} is a unified unsupervised framework for incident management in cloud systems. It integrates multimodal observations and applies representation learning techniques to perform anomaly detection without requiring labeled anomaly samples.
\end{itemize}

\subsubsection{Evaluation Metrics}

We adopt Precision, Recall, and F1 score as evaluation metrics. Since anomalies in cloud systems typically manifest as continuous segments rather than isolated time points, we perform evaluation at the segment level. Specifically, if any point within an anomalous segment is correctly detected, the entire segment is considered a true positive. Conversely, if no point within an anomalous segment is detected, the entire segment is counted as a false negative. For normal segments, if any point is incorrectly flagged as anomalous, it is treated as a false positive. Based on these segment-level criteria, Precision measures the fraction of detected anomalous segments that are truly anomalous, Recall measures the fraction of actual anomalous segments that are successfully detected, and F1 score is the harmonic mean of Precision and Recall, providing a balanced assessment of overall detection performance.

\subsubsection{Implementation Details} 
All experiments are conducted using PyTorch~\cite{paszke2019pytorch} in Python 3.10 and executed on an NVIDIA Tesla-A800 GPU. The model is optimized using Adam~\cite{kingma2014adam} with an initial learning rate of 0.001 and a batch size of 32. The random seed is set to 2026. The warm-up stage is trained for 3 epochs using the reconstruction loss~\eqref{eq25} and the IB-select loss~\eqref{eq11} to stabilize the multimodal fusion space. After prototype initialization, the model enters the joint optimization stage and is further trained for 50 epochs. During training, the learning rate is decayed using the StepLR scheduler. The textual embedding model is the pre-trained model ``text-embedding-v4''~\cite{openai2024embeddings}. The time window length is set to 60, the number of prototypes is set to 15, and the number of scales is set to 3. 

\begin{table*}[t]
\centering
\caption{Overall anomaly detection results. The best-performing results are highlighted in bold, and the second-best results are highlighted in underline.}
\label{tab: main}
\begin{tabular}{c|ccc|ccc|ccc|ccc}
\toprule
 & \multicolumn{3}{c|}{\textbf{Modality}} & \multicolumn{3}{c|}{\textbf{AIOps2021}} & \multicolumn{3}{c|}{\textbf{Train-Ticket}} & \multicolumn{3}{c}{\textbf{D1}} \\ \cmidrule{2-13} 
\multirow{-3}{*}{\textbf{Method}} & Metric & Log & Topology & Precision & Recall & F1 & Precision & Recall & F1 & Precision & Recall & F1 \\
\midrule
ContraAD~\cite{ContraAD} & \cmark &  &  & 0.6527 & 0.5792 & 0.6138 & 0.5146 & 0.5282 & 0.5210 & 0.5424 & 0.5731 & 0.5573 \\
TranAD~\cite{Tuli2022tranad} & \cmark &  &  & 0.6281 & 0.5549 & 0.5892 & 0.4917 & 0.5378 & 0.5137 & 0.5179 & 0.5553 & 0.5359 \\
Hades~\cite{lee2023hase} &  & \cmark &  & 0.5138 & 0.4693 & 0.4905 & 0.4615 & 0.5051 & 0.4823 & 0.3671 & 0.4002 & 0.3829 \\
MultiLog~\cite{zhang2024multilog} &  & \cmark &  & 0.4786 & 0.4439 & 0.4606 & 0.4937 & 0.4829 & 0.4882 & 0.2576 & 0.3357 & 0.2915 \\
\midrule
DiagFusion~\cite{DiagFusion} & \cmark & \cmark &  & 0.2341 & 0.1655 & 0.1969 & 0.1123 & 0.1408 & 0.1249 & 0.3446 & 0.2891 & 0.3144 \\
Medicine~\cite{Medicine} & \cmark & \cmark &  & 0.6829 & 0.1761 & 0.2800 & 0.5012 & 0.2198 & 0.3056 & 0.6194 & 0.4576 & 0.5263 \\
FAMOS~\cite{famos} & \cmark & \cmark &  & 0.5235 & 0.5811 & 0.5508 & 0.5732 & \uline{0.5813} & \uline{0.5772} & 0.7654 & \uline{0.8552} & \uline{0.8078} \\
Eadro~\cite{eadro} & \cmark & \cmark & \cmark & 0.6778 & \uline{0.5912} & \uline{0.6315} & \uline{0.5893} & 0.5554 & 0.5718 & \uline{0.7835} & 0.7529 & 0.7679 \\
ART~\cite{art} & \cmark & \cmark & \cmark & \uline{0.6883} & 0.3333 & 0.4491 & 0.5677 & 0.4207 & 0.4833 & 0.7092 & 0.4011 & 0.5124  \\
\midrule
\textbf{BEACON} & \cmark & \cmark & \cmark & \textbf{0.7368} & \textbf{0.6123} & \textbf{0.6688} & \textbf{0.6311} & \textbf{0.6274} & \textbf{0.6292} & \textbf{0.8562} & \textbf{0.8890} & \textbf{0.8723} \\
\bottomrule
\end{tabular}
\end{table*}

\subsection{Main Results}
Table~\ref{tab: main} presents the overall performance of all methods on the three datasets. \textbf{BEACON} achieves the strongest results among the compared methods on these datasets. To verify whether the improvements over the baselines are statistically significant, we conduct t-tests comparing the anomaly scores of our method and the baselines. The resulting p-values are smaller than 0.001, indicating that the improvements are statistically significant.

\subsubsection{Compared to single-modal methods}
For single-modal methods, ContraAD and TranAD rely on metric data, while Hades and MultiLog focus on log information. Overall, metric-based methods perform better than log-based approaches on all datasets. For example, ContraAD achieves an F1 score of 0.6271 on AIOps2021, which is higher than Hades and MultiLog. This suggests that system metrics provide relatively stable information for anomaly detection. However, the overall performance of single-modal methods remains limited. Even the best metric-based baseline, ContraAD, achieves only moderate F1 scores on the three datasets. Log-based methods show even larger performance degradation, especially on the D1 dataset, indicating that relying solely on log information is insufficient in complex cloud environments.

In comparison, \textbf{BEACON} consistently outperforms all single-modal baselines. 
Specifically, it achieves a relative average improvement of 28.7\% in F1 score over the strongest single-modal method when evaluated on the three datasets. The results indicate that integrating multiple modalities helps capture complementary anomaly information that is difficult to observe from a single data source.

\subsubsection{Compared to multimodal methods}
Multimodal methods generally achieve better performance than single-modal approaches, showing that combining metrics, logs, and topology provides richer information for anomaly detection. However, their effectiveness largely depends on the representation and integration of heterogeneous modalities.

Some methods exhibit clear limitations in modeling multimodal observations. For example, DiagFusion transforms multimodal observations into textual event sequences and applies language modeling techniques. This representation simplifies diverse modalities into discrete events and may discard important temporal and numerical characteristics, leading to significant performance degradation. Medicine attempts to balance the influence of different modalities and achieves relatively high precision on some datasets, but its recall remains very low. This suggests that overly constraining modality contributions may make the model conservative and cause many anomalies to be missed. In addition, early fusion strategies such as ART directly concatenate multimodal representations without explicitly addressing modality heterogeneity, which limits the effectiveness of the fused representation. More recent approaches attempt to model interactions across modalities. Methods such as Eadro and FAMOS employ graph attention or cross-attention mechanisms and achieve relatively strong results across datasets. Their performance indicates that properly modeling dependencies among multimodal representations is beneficial for anomaly detection.

As shown in Table~\ref{tab: main}, \textbf{BEACON} consistently achieves the best performance across all datasets. Compared with the strongest baselines, \textbf{BEACON} improves the F1 score by 5.9\% on AIOps2021 and 9.0\% on Train-Ticket, and achieves a 7.9\% higher F1 score than the best baseline on the D1 dataset.
These results suggest that fine-grained semantic fusion, achieved by aligning multimodal representations in a shared space, helps improve the model’s anomaly detection capability.

\subsection{Ablation Studies}

To evaluate the contribution of each component in BEACON, we design several variants and report the results in Table~\ref{tab: ablation}.

\begin{table}[b]
\centering
\caption{Effect of \textbf{BEACON} variants on F1 across all datasets.}
\label{tab: ablation}
\begin{tabular}{c|c|c|c}
\toprule
\textbf{Method} & \textbf{AIOps2021} & \textbf{Train-Ticket} & \textbf{D1} \\
\midrule
w/o intrinsic semantics & 0.6276 & 0.6004 & 0.8577 \\
w/o temporal dynamics & 0.6154 & 0.5913 & 0.8516 \\
w/o prototypes & 0.6253 & 0.5998 & 0.8572 \\
w/o multi-scale attention & 0.6227 & 0.6039 & 0.8468 \\ 
w/o topology & 0.6410 & 0.6156 & 0.8614 \\
w/o warm-up & 0.6578 & 0.6198 & 0.8537 \\
\midrule
\textbf{BEACON} & \textbf{0.6688} & \textbf{0.6292} & \textbf{0.8723} \\
\bottomrule
\end{tabular}
\end{table}

\subsubsection{w/o intrinsic semantics}
discards the KL-divergence regularization term in~\eqref{eq7} of the IB-guided dynamic metric selector.
Intrinsic semantics provide stable descriptions of metric meanings, which help the selector distinguish informative metrics from noisy ones. Without this information, the selector relies mainly on observed signals, which makes metric selection less reliable and reduces overall performance.

\subsubsection{w/o temporal dynamics}
This variant excludes the dynamic temporal features defined in~\eqref{eq8} from the IB-guided dynamic metric selector, retaining only static semantic representations.
The results show a noticeable performance drop than removing intrinsic semantics. This suggests that static semantics overlook state-dependent metric relevance, whereas modeling temporal dynamics enables DIMA's adaptive retrieval.

\subsubsection{w/o prototypes}
This variant removes the semantic prototype projection and directly fuses modality features through cross-attention. 
Prototypes serve as an intermediate semantic space for modality alignment, projecting heterogeneous features into a shared representation space. Without this step, cross-attention operates on modality representations distributed in different feature spaces, which weakens cross-modality alignment and reduces the effectiveness of feature fusion.

\subsubsection{w/o multi-scale attention}
This variant replaces the multi-scale temporal delay kernels with a standard transformer. 
It results in a noticeable performance drop, demonstrating that anomaly propagation in real systems often involves different temporal delays. Modeling multiple temporal scales is important for capturing these asynchronous dependencies.

\subsubsection{w/o topology}
This variant removes topology-aware spatial aggregation and models each node independently. 
The results show a slight performance decrease after removing spatial modeling. A possible reason is that the multi-scale temporal attention captures cross-instance temporal dependencies over different time ranges, which reflects the spatial relationships among instances. Thus, removing topology-aware modeling has a relatively limited impact on the overall performance.

\subsubsection{w/o warm-up}
This variant removes the warm-up training stage for prototype initialization and directly trains the full model end-to-end from scratch. 
The warm-up stage stabilizes the initialization of semantic prototypes. Without this stage, prototypes are randomly initialized, which leads to unstable prototype representations in early training. This weakens their ability to guide modality alignment and feature fusion, leading to reduced performance.

\subsection{Parameter Sensitivity Analysis}

We evaluate the robustness of \textbf{BEACON} to two key hyperparameters: the number of prototypes $K$ in DIMA and temporal scales $M$ in ASTP. We vary one parameter at a time while fixing the other, reporting F1 scores across all datasets.

\subsubsection{Effect of the number of prototypes $K$} 
\label{parameter:p}
\begin{figure}[h]
    \centering
    \begin{minipage}[t]{0.45\textwidth}
        \centering
        \includegraphics[width=\textwidth]{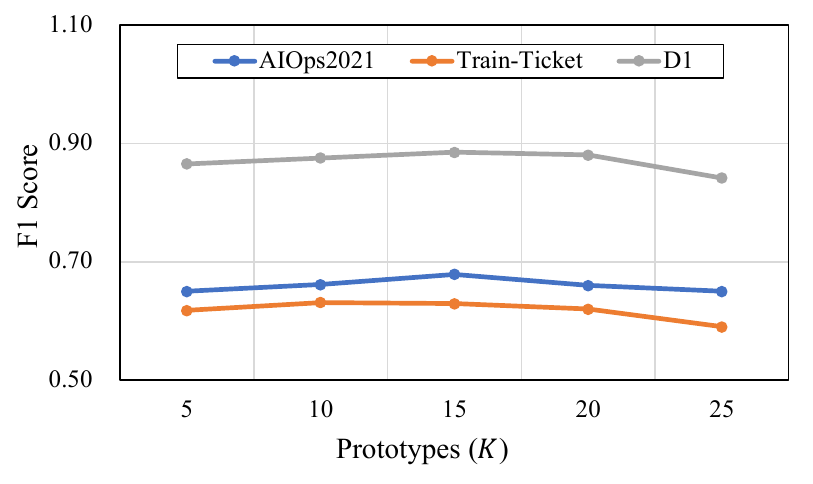}
        \subcaption{Effect of the number of prototypes $K$.}
        \label{fig:prototypes}
    \end{minipage}
    \hfill
    \begin{minipage}[t]{0.45\textwidth}
        \centering
        \includegraphics[width=\textwidth]{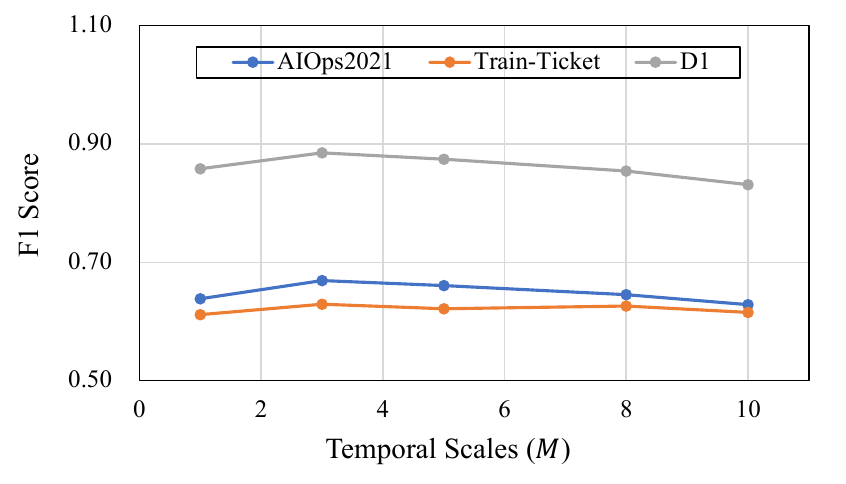}
        \subcaption{Effect of the number of temporal scales $M$.}
        \label{fig:scales}
    \end{minipage}
    
    \caption{Results of \textbf{BEACON} with different parameters on F1 score across three datasets.}
    \label{fig:pm}
\end{figure}

To study the impact of the number of prototypes, we set $K$ to \{5, 10, 15, 20, 25\} and report the results in Fig.~\ref{fig:pm}(a). The best performance on AIOps2021 and D1 is achieved around $K=15$, while Train-Ticket reaches its highest F1 score at $K=10$. A possible reason is that Train-Ticket is collected with a relatively dense sampling interval (5 seconds), where system states change smoothly between adjacent observations. As a result, a smaller number of prototypes is sufficient to capture the main patterns. Considering that all datasets achieve strong performance around $K=15$, we set $K=15$ as the default configuration in the experiments. As $K$ increases, the F1 score first improves and then slightly declines across all datasets. This trend indicates that an appropriate number of prototypes helps capture diverse patterns in multimodal observations and improves representation learning for anomaly detection, while too many prototypes may introduce redundant patterns and degrade performance. Therefore, a moderate number of prototypes provides a good balance between representation capacity and model robustness.

\subsubsection{Effect of the number of temporal scales $M$} 
To study the impact of the number of temporal scales, we set $M$ to \{1, 3, 5, 8, 10\} and report the results in Fig.~\ref{fig:pm}(b). As $M$ increases from 1 to 3, the F1 score on all three datasets improves, indicating that modeling multiple temporal ranges helps capture asynchronous temporal dependencies. The best performance is achieved around $M=3$, suggesting that the model can effectively capture short-, medium-, and long-term temporal dependencies with a moderate number of temporal scales. This design enables the model to adaptively aggregate temporal information at different ranges and improves anomaly detection performance. When $M$ further increases, the performance slightly declines, likely due to redundant temporal information and increased model complexity. In contrast, the performance on the Train-Ticket dataset changes only marginally as $M$ increases. This is consistent with the observation in Section~\ref{parameter:p}, likely because the dataset is collected with a fine-grained sampling interval, where observations are densely sampled, and system states exhibit limited variation within the same temporal windows.

\begin{figure*}[t]
  \centering
  \includegraphics[width=1\linewidth]{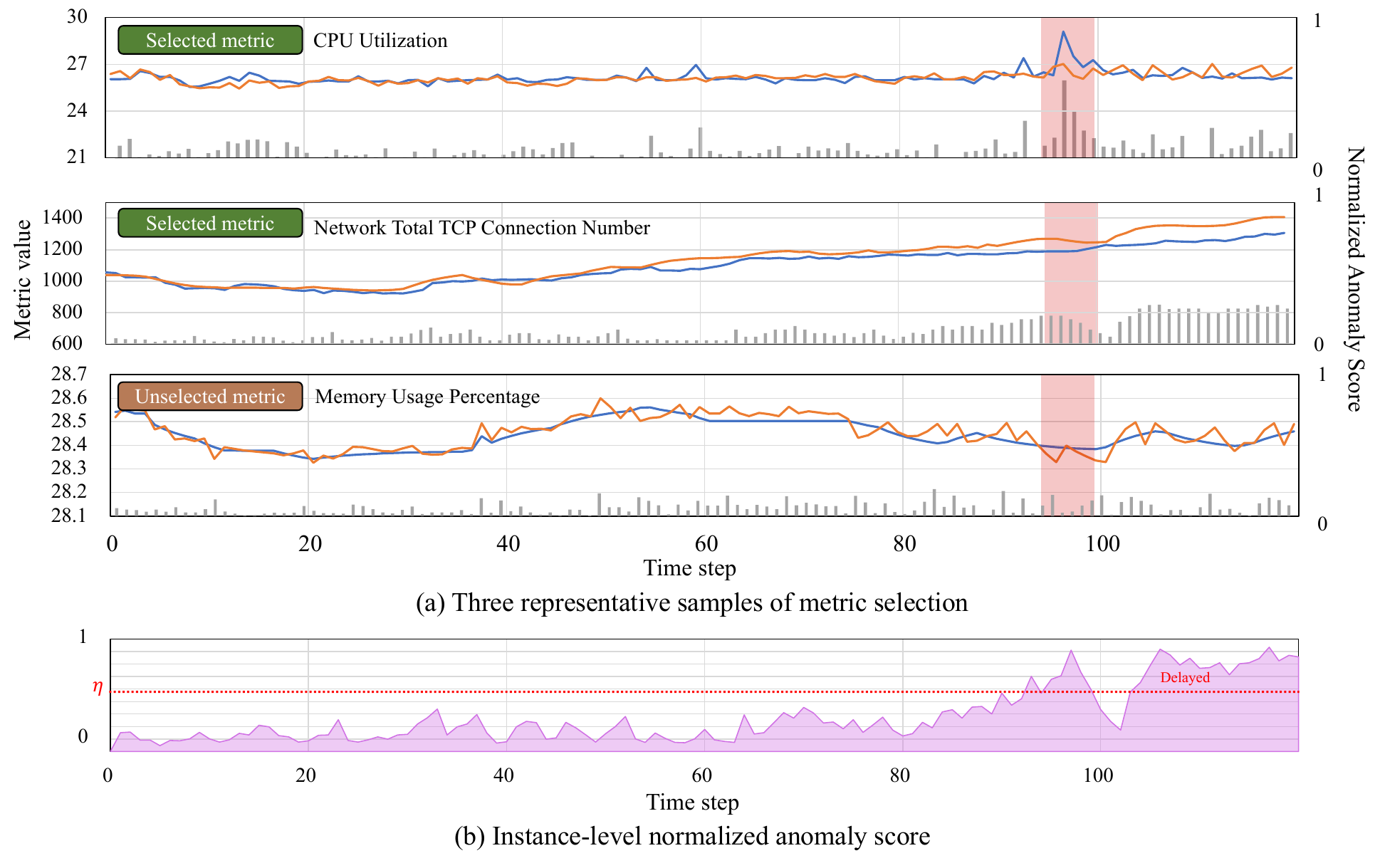}
  \caption{Visualization of metric selection and instance-level anomaly detection. (a) Three representative metrics from the same instance. Blue lines and orange lines denote observed and reconstructed values, respectively. Gray bars represent normalized anomaly scores, and the red shaded region indicates the ground-truth anomaly interval. (b) The aggregated instance-level normalized anomaly score, $\eta$ is the anomaly threshold on the validation set.}
  \label{fig: case}
\end{figure*}

\subsection{Efficiency}
\begin{figure}[h]
  \centering
  \includegraphics[width=1\linewidth]{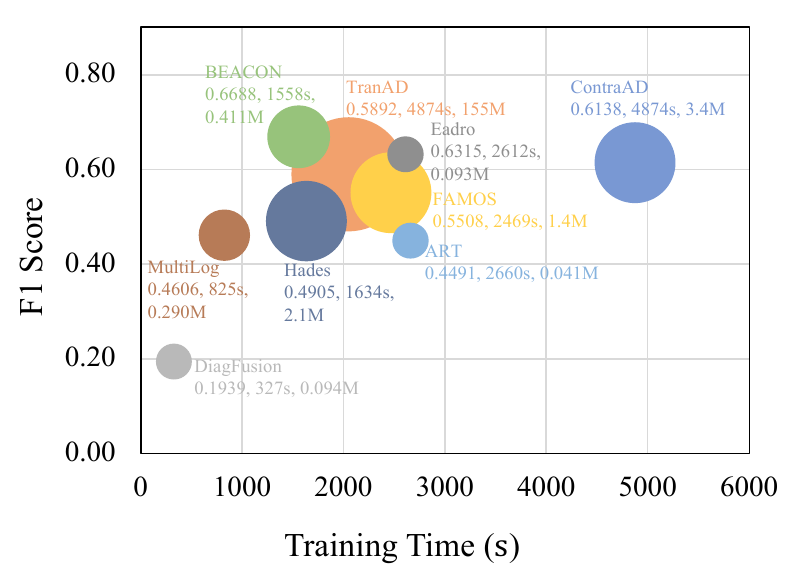}
  \caption{Comparison of F1 Score, training time, and parameter sizes on the AIOps2021 dataset.}
  \label{fig: time}
\end{figure}

Fig.~\ref{fig: time} compares different methods in terms of F1 score, training time, and parameter size. The results show clear differences in the trade-off between detection performance, computational cost, and model complexity.

\textbf{BEACON} achieves the best overall detection performance while maintaining moderate training time and a relatively small parameter size. Some baselines obtain competitive F1 scores but rely on significantly larger models or longer training time, which increases the computational overhead. Several lightweight approaches reduce parameter size or training time, but their detection performance is noticeably lower. This suggests that simply reducing model complexity often leads to a loss in accuracy. In contrast, \textbf{BEACON} maintains a compact model while still achieving strong performance. 
Overall, the results indicate that \textbf{BEACON} achieves a better balance between effectiveness and efficiency. 

\subsection{Case Studies}
To demonstrate the effectiveness of the main components in \textbf{BEACON}, which identify informative metrics from massive multimodal observations for fusion with logs and effectively model asynchronous spatio-temporal dependencies, we provide a ``\textit{Network Packet Loss}'' anomaly sample in Fig.~\ref{fig: case}. 

We observe that ``\textit{Network Packet Loss}'' triggers request retries and connection timeouts, resulting in a significant increase in ``\textit{CPU utilization}''. By combining the temporal pattern of ``\textit{CPU Utilization}'' with its semantic meaning conveyed by the metric name, the aggregator correctly identifies it as an anomaly-relevant metric. 
Notably, the large number of repeated requests also leads to a delayed increase in ``\textit{Network Total TCP Connection Number}''. This delayed change is also captured by the model, and the corresponding metric is correctly identified as anomaly-related. 
Similarly, ``\textit{Memory Usage Percentage}'' shows no significant variation in its temporal pattern and is excluded. 
Fig.~\ref{fig: case}(b), the instance-level anomaly score remains low during normal periods and rises sharply within the ground-truth anomaly interval, and anomalies caused by delayed effects are also correctly identified. This is because the aggregator assigns high weights to the selected anomaly-relevant metrics while suppressing the influence of irrelevant ones, allowing the aggregated score to faithfully reflect the actual anomaly occurrence without being diluted by noise.

This case illustrates that the DIMA can effectively distinguish anomaly-relevant metrics from irrelevant ones. The joint use of temporal dynamics and metric semantics enables more precise metric selection than relying on either modality alone, reducing false positives caused by coincidental fluctuations and improving the interpretability of anomaly detection results. Meanwhile, the ASTP captures asynchronous spatio-temporal dependencies among system instances through explicit multi-scale temporal modeling and spatial graph learning.

\section{Related Work}

\subsection{Anomaly Detection}

Traditional anomaly detection methods aim to identify observations that deviate from the expected data generation process in time series. Early approaches mainly rely on classical statistical and signal processing techniques. Representative methods include density- and clustering-based approaches such as LOF~\cite{breunig2000lof} and CBLOF~\cite{he2003discovering}, and histogram-based methods like HBOS~\cite{goldstein2012histogram}. Other methods use statistical thresholds or signal transformations, including z-score-based detection, matrix profile, and discrete wavelet transform~\cite{bhatnagar2021merlion,thill2017time,yeh2016matrix,BoniolP20}. These approaches typically detect anomalies based on statistical deviation or pattern mismatch in time series, struggling to capture complex system behaviors and high-dimensional monitoring signals in modern cloud environments. 

With the advancement of deep learning techniques, numerous methods have been proposed to model single-modal observability data. Some methods adopt autoencoders and transformers to detect deviations in multivariate time series~\cite{ContraAD,Tuli2022tranad,ChenL00LL22,RenXWYHKXYTZ19,li2021multivariate,li2025tsfm,AudibertMGMZ20, ChenHCCRSYPG25, zhang26icde}, while others parse logs into event sequences and perform anomaly detection based on sequential patterns~\cite{JiaLYHW22,ccs,LogTransfer,zhang2024multilog,lee2023hase}. Although these single-modal approaches achieve promising results within their respective data sources, they inherently miss anomaly signals that only manifest in other modalities, limiting their overall detection capability. 

Recently, integrating multimodal monitoring data has attracted increasing attention in anomaly detection~\cite{LeeYCSYL23, Zhang0SZFWL022, Identifying,mindts,TimeMMD,calf}. 
Some methods encode each modality independently and perform early fusion by concatenating embeddings~\cite{art}, while others apply late fusion to combine modality-specific representations~\cite{Medicine}. Cross-modal graph structures have also been used to model correlations across modalities~\cite{eadro}. However, these approaches often fail to capture the heterogeneity across modalities. Although some multimodal time series models attempt to align time series with text~\cite{mindts}, they assume the availability of paired textual descriptions and are not directly applicable to cloud system scenarios where metrics and logs exhibit distinct structures and semantics. In contrast, DIMA filters redundant information through an IB mechanism and aligns multimodal representations in a shared prototype space.

\subsection{Spatio-Temporal Modeling}

Spatio-temporal modeling aims to capture both temporal dynamics and spatial dependencies in data. 
Early efforts primarily modeled the temporal dynamics of time series such as RNNs, temporal convolution networks (TCNs), and Transformers, which capture temporal correlations in system behaviors~\cite{AutoCTS,SSDts}.  
Methods such as PatchTST~\cite{patchtst} and TriFormer~\cite{cirstea2022triformer} learn temporal dependencies by modeling local temporal patterns within segmented sequences using Transformer architectures. Other approaches, including TimesNet~\cite{timesnet} and TFMixer~\cite{qiu2026bridging}, capture temporal dynamics through frequency-domain decomposition, such as FFT or Wavelet Transform, to model multi-scale temporal patterns. While effective for temporal modeling, these approaches treat system entities independently and ignore structural relationships among instances.

To address this limitation, some studies introduce graph-based models that incorporate spatial dependencies\cite{cirstea2021graphattentionrecurrentneural,tian2024air-dualode,fang2024efficient,tian2025arrow,xu2024mmpath,mei2025find3,PCTP}. 
For example, Orca~\cite{li2024cikm} leverages Z-order curves to capture grid-based spatial proximity, while MOST~\cite{xu2026most} employs a multimodal guided spatial aggregation module with a mixture-of-experts architecture to model spatio-temporal relationships among neighboring sensors. In cloud systems, spatial relationships among instances are reflected through service traces and system dependency graphs. Several methods detect system anomalies by identifying sudden increases in latency from trace data~\cite{Medicine}, while others model the spatial propagation of anomalies based on classical graph algorithms such as cut trees~\cite{art} and random walks~\cite{ClusterRCA}. However, these methods assume that anomalies manifest simultaneously across related instances, overlooking the fact that in complex cloud systems, anomalies often propagate as cascading failures with varying temporal delays. In contrast, the proposed ASTP module models asynchronous spatio-temporal dependencies across instances through learnable multi-scale delay kernels and graph learning.

\section{Conclusion}
We introduce BEACON, a multimodal anomaly detection framework for cloud systems that integrates time series metrics, unstructured or semi-structured logs, and system dependency graphs. 
\textbf{BEACON} addresses both challenges through two core components. 
The Dynamic Interpretable Multimodal Aggregator (DIMA) compresses metric observations via an information bottleneck mechanism and aligns the informative metric embeddings with log embeddings in a shared prototype space for effective multimodal fusion. The Asynchronous Spatio-Temporal Propagator (ASTP) further models dependencies across temporal dynamics and system structure through multi-scale temporal modeling and topology-aware aggregation, allowing anomaly signals to propagate across instances and time. Experiments on three real-world cloud datasets show that \textbf{BEACON} consistently outperforms existing methods in anomaly detection. These results highlight the effectiveness of fine-grained multimodal semantic fusion and modeling asynchronous spatio-temporal dependencies in complex cloud environments.

\section*{Acknowledgment}
This work was partially supported by the National Natural Science Foundation of China (62406112) and Alibaba Innovative Research Program.

\clearpage

\bibliographystyle{IEEEtran}
\bibliography{IEEEabrv,reference}

\end{document}